\pdfoutput=1

\documentclass[11pt]{article}

\usepackage{emnlp2021}

\usepackage{times}
\usepackage{latexsym}

\usepackage[T1]{fontenc}

\usepackage[utf8]{inputenc}

\usepackage{microtype}

\usepackage{booktabs}
\usepackage{multirow}
\usepackage{graphicx}
\usepackage{pifont}
\usepackage{amsmath}
\usepackage{amssymb}
\usepackage{ifthen}
\usepackage{dsfont}

\usepackage{tikz}
\usetikzlibrary{bayesnet}
\usepackage{subfloat}
\usepackage{subfig}
\usepackage[linguistics]{forest}

\usepackage[ruled,vlined]{algorithm2e}

\newcommand{\prob}[2][]{\text{\bf P}\ifthenelse{\not\equal{}{#1}}{_{#1}}{}\!\left(#2\right)}
\newcommand{\expect}[2][]{\text{\bf E}\ifthenelse{\not\equal{}{#1}}{_{#1}}{}\!\left[#2\right]}
\newcommand{\var}[2][]{\text{\bf Var}\ifthenelse{\not\equal{}{#1}}{_{#1}}{}\!\left[#2\right]}

\newcommand{\smatch}{\textsc{Smatch}}

\newcommand{\shift}{\textsc{shift}}

\newcommand{\cp}{\textsc{copy}}

\newcommand{\la}[1]{\textsc{la(#1)}}
\newcommand{\ra}[1]{\textsc{ra(#1)}}

\title{Structure-aware Fine-tuning of Sequence-to-sequence Transformers for Transition-based AMR Parsing}

\author{Jiawei Zhou$^\text{\ding{171}}$ \ \ \ Tahira Naseem$^\Diamond$ \ \ \ Ram\'{o}n Fernandez Astudillo$^\Diamond$ \ \ \ Young-Suk Lee$^\Diamond$ \\ \textbf{Radu Florian$^\Diamond$ \ \ \  Salim Roukos$^\Diamond$} \\
        $^\text{\ding{171}}$Harvard University \hspace{1cm} $^\Diamond$IBM Research \\ 
        $^\text{\ding{171}}$\texttt{\small{jzhou02@g.harvard.edu}} \ \ \
        $^\Diamond$\texttt{\small{ramon.astudillo@ibm.com}} \\
        $^\Diamond$\texttt{\small{\{tnaseem, ysuklee, raduf, roukos\}@us.ibm.com}}
        }

\begin{document}
\maketitle
\begin{abstract}
Predicting linearized Abstract Meaning Representation (AMR) graphs using pre-trained sequence-to-sequence Transformer models has recently led to large improvements on AMR parsing benchmarks. These parsers are simple and avoid explicit modeling of structure but lack desirable properties such as graph well-formedness guarantees or built-in graph-sentence alignments. In this work we explore the integration of general pre-trained sequence-to-sequence language models and a structure-aware transition-based approach.  
We depart from a pointer-based transition system and propose a simplified transition set, designed to better exploit pre-trained language models for structured fine-tuning. We also explore modeling the parser state within the pre-trained encoder-decoder architecture and different vocabulary strategies for the same purpose. 
We provide a detailed comparison with recent progress in AMR parsing and show that the proposed parser retains the desirable properties of previous transition-based approaches, while being simpler and reaching the new parsing state of the art for AMR 2.0, without the need for graph re-categorization.
\end{abstract}

\section{Introduction}
\label{sec:intro}






The task of Abstract Meaning Representation (AMR) parsing translates a natural sentence into a rooted directed acyclic graph capturing the semantics of the sentence, with nodes representing concepts and edges representing their relations \citep{banarescu2013abstract}.
Recent works utilizing pre-trained encoder-decoder language models show great improvements in AMR parsing results \citep{xu2020improving, bevilacqua2021one}. These approaches avoid explicit modeling of the graph structure. Instead, they directly predict the linearized AMR graph treated as free text. While the use of pre-trained Transformer encoders is widely extended in AMR parsing, the use of pre-trained Transformer decoders is recent and has shown to be very effective, maintaining current state-of-the-art results \citep{bevilacqua2021one}.

These approaches however lack certain desirable properties. There are no structural guarantees of graph well-formedness, i.e. the model may predict strings that can not be decoded into valid graphs, and post-processing is required. Furthermore, predicting AMR linearizations ignores the implicit alignments between graph nodes and words, which provide a strong inductive bias and are useful for downstream AMR applications \citep{mitra2016addressing, liu2018toward, vlachos2018guided,kapanipathi-etal-2021-leveraging,naseem2021semantics}.

On the other hand, transition-based AMR parsers \citep{wang2015transition, ballesteros-al-onaizan-2017-amr, astudillo2020transition, zhou2021amr} operate over the tokens of the input sentence, generating the graph incrementally. They implicitly model graph structural constraints through transitions and yield alignments by construction, thus guaranteeing graph well-formedness.\footnotemark\footnotetext{With the only exception being disconnected graphs, which happen infrequently in practice.} 
However, it remains unclear whether explicit modeling of structure is
still beneficial for AMR parsing in the presence of powerful pre-trained language models and their strong free text generation abilities.

In this work, we integrate pre-trained sequence-to-sequence (seq-to-seq) language models with the transition-based approach for AMR parsing, and explore to what degree they are complementary.
To fully utilize the generation power of the pre-trained language models, we propose a transition system with a small set of basic actions -- a generalization of the action-pointer transition system of \citet{zhou2021amr}. 
We use BART  \citep{lewis2019bart} as our pre-trained language model, since it has shown significant improvements in linearized AMR generation \citep{bevilacqua2021one}. Unlike previous approaches that directly fine-tune the model with linearized graphs, we modify the model structure to work with our transition system, and encode parser states in BART's attention mechanism \citep{astudillo2020transition, zhou2021amr}. We also explore different vocabulary strategies for action generation. These changes convert the pre-trained BART to a transition-based parser where graph constraints and alignments are internalized.

We provide a detailed comparison with top-performing AMR parsers and perform ablation experiments showing that our proposed transition system and BART modifications are both necessary to achieve strong performance.
Although BART has great language generation capacity, it still benefits from parser state encoding with hard attention, and can efficiently learn structural output. Our model establishes a new state of the art for AMR 2.0 while maintaining graph well-formedness guarantees and producing built-in alignments.

\section{Intricacies of AMR Parsers}
\label{sec:complexity}


A frequent complaint about AMR parsers is that they involve combining many different techniques and hand-crafted rules, resulting in complex pipelines that are hard to analyze and generalize poorly. This situation has notably improved in the past few years but there are still two main sources of complexity present in almost all recent parsers: graph re-categorization and subgraph actions.

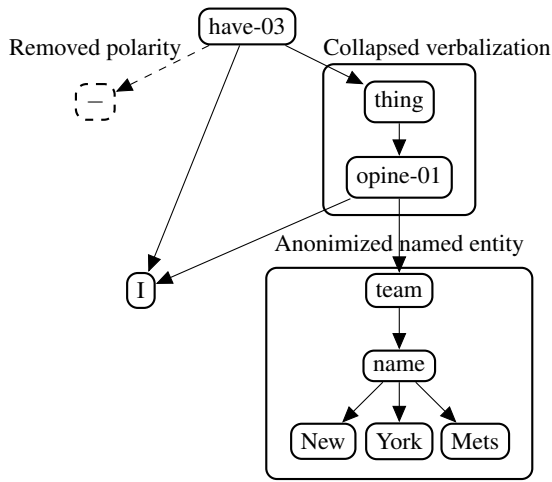
\begin{figure}[!t]
    \centering
\hspace{-1cm}
\begin{tikzpicture}

\node [draw,rounded corners,thick] (have) at (3, 6) {\footnotesize have-03};

\node [draw,rounded corners,thick,dashed] (minus) at (1, 5) {\footnotesize $-$};
\draw [->,dashed] (have) to (minus);
\node [] (mark0) at (1, 5.7) {\footnotesize Removed polarity};

\node (rect) at (5, 4.5) [draw,rounded corners,thick,minimum width=20mm,minimum height=20mm] {};
\node [draw,rounded corners,thick] (thing) at (5,5) {\footnotesize thing};
\node [draw,rounded corners,thick] (opine) at (5,4) {\footnotesize opine-01};
\draw [->] (have) to (thing);
\draw [->] (thing) to (opine);
\node [] (mark1) at (5.5, 5.7) {\footnotesize Collapsed verbalization};

\def \ney {2.5}
\node (rect) at (5, \ney - 1.1) [draw,rounded corners,thick,minimum width=35mm,minimum height=28mm] {};
\node [draw,rounded corners,thick] (team) at (5,\ney) {\footnotesize team};
\node [draw,rounded corners,thick] (name) at (5,{\ney - 1}) {\footnotesize name};
\node [draw,rounded corners,thick] (new) at (4,{\ney - 2}) {\footnotesize New};
\node [draw,rounded corners,thick] (york) at (5,{\ney - 2}) {\footnotesize York};
\node [draw,rounded corners,thick] (mets) at (6,{\ney - 2}) {\footnotesize Mets};
\draw [->] (opine) to (team);
\draw [->] (team) to (name);
\draw [->] (name) to (new);
\draw [->] (name) to (york);
\draw [->] (name) to (mets);
\node [] (mark2) at (5, \ney + 0.6) {\footnotesize Anonimized named entity};

\node [draw,rounded corners,thick] (i) at (1.6, \ney) {\footnotesize I};
\draw [->] (have) to (i);
\draw [->] (opine) to (i);


%
%
%
\vspace{1cm}

\end{tikzpicture}

\caption{AMR graph of the sentence \textit{I have no opinion on the New York Mets}. Examples of subgraphs for entity anonymization, collapsing of verbalized nouns and removal of the polarity node and edge.}
\label{fig:graph_embedding}
\end{figure}

Graph re-categorization \citep{wang2017getting,lyu2018amr,zhang2019amr} normalizes the graph prior to learning. This includes joining certain subgraphs such as entities, dates and other constructs into single nodes, removing special types of nodes like polarity and normalizing propbank names. An example of common normalizations is displayed in Figure~\ref{fig:graph_embedding}. Training and decoding of models using this technique happens in this re-categorized space. Re-categorized graphs are expanded to normal valid AMR graphs in a post-processing stage. The type and number of subgraphs normalized vary across implementations, but most high performing approaches \citep{cai2020amr,bevilacqua2021one} utilize the re-categorization described in Appendix A.1 of \citet{zhang2019amr}. This version requires of an external Named Entity Recognition (NER) system to anonymize named entities, both at train time and test time. It also makes use of look-up tables for nominalizations  (e.g. \textit{English} to \textit{England}) and other hand-crafted rules. Graph re-categorization has been criticised for its lack of generalization to new domains, such as biomedical domain or even the AMR 3.0 corpus \citep{bevilacqua2021one}. Recent top performing systems e.g. \citet{cai2020amr,bevilacqua2021one} also provide results without re-categorization, but this is shown to hurt performance notably on the AMR 2.0 corpus.

Subgraph actions \citep{ballesteros2017amr} are used in transition-based systems and play a role similar to re-categorization. Instead of normalizing and reverting, transition-based parsers apply a subgraph action that generates an entire subgraph at once. This subgraph action coincides with many of the subgraphs collapsed in re-categorization. Subgraph actions bring however fewer external dependencies, since the parser learns to segment and identify subgraphs during training. They still suffer however from data sparsity since some subgraphs appear very few times. As in re-categorization, subgraph actions also make use of lookup tables for nominalization and similar constructs that hinder generalization. Furthermore, they create the problem of unattachable nodes. This was addressed in \citet{zhou2021amr} by ignoring subgraphs for a set of heuristically determined cases. Subgraph actions have been used in all transition-based AMR systems \citep{naseem2019rewarding,astudillo2020transition,zhou2021amr}.

Aside from NER, past AMR parsers have other external dependencies such as POS taggers \cite{zhang2019amr,cai2020amr} and lemmatizers \cite{cai2020amr, naseem2019rewarding,astudillo2020transition}. 

\begin{figure*}
    \centering
\begin{tikzpicture}


\def \xwarp {1.1}
\def \yposa {1}

\def \xpos {1}
\node [] (action) at ({\xpos * \xwarp}, 0.5) {\footnotesize 1};
\node [] (action) at ({\xpos * \xwarp}, 0) {\footnotesize person};
\node [] (token) at ({\xpos * \xwarp},1) {\footnotesize Employees};
\node [draw,rounded corners,thick] (employee01) at ({\xpos * \xwarp}, 2) {\footnotesize employ-01};
\path (token) edge[-,draw,dashed,thick] (employee01);
\node [draw,rounded corners,thick] (person) at ({\xpos * \xwarp}, 3) {\footnotesize person};
\draw [latex-] (employee01) to node[right] {{\footnotesize \textsc{arg1\textup{-of}} }} (person);

\def \xpos {2.4}
\node [] (action) at ({\xpos * \xwarp}, 0.5) {\footnotesize 2};
\node [] (action) at ({\xpos * \xwarp}, 0) {\footnotesize employ-01};
\node [] (token) at ({\xpos * \xwarp},1) [text=gray] {\footnotesize Employees};
\node [] (arc\xpos) at ({\xpos * \xwarp}, -0.5) {\footnotesize $\textsc{ra(1,arg1-\textup{of})}$};

\def \xpos {3.8}
\node [] (token) at ({\xpos * \xwarp},1) [text=gray] {\footnotesize Employees};
\node [] (action) at ({\xpos * \xwarp}, 0) {\footnotesize $\textsc{shift}$};

\def \xpos {4.9}
\node [] (action) at ({\xpos * \xwarp}, 0.5) {\footnotesize 5};
\node [] (action) at ({\xpos * \xwarp}, 0) {\footnotesize like-01};
\node [] (token) at ({\xpos * \xwarp},1) {\footnotesize liked};
\node [draw,rounded corners,thick] (like01) at ({\xpos * \xwarp}, 4) {\footnotesize like-01};
\path (token) edge[-,draw,dashed,thick] (like01);
\draw [latex-] (person) to node[right] {{\footnotesize \textsc{arg0}}} (like01);
\node [] (arc\xpos) at ({\xpos * \xwarp}, -0.5) {\footnotesize $\textsc{la(1,arg0)}$};

\def \xpos {5.8}
\node [] (action) at ({\xpos * \xwarp}, 0) {\footnotesize \textsc{root}};
\node [] (token) at ({\xpos * \xwarp},1) [text=gray] {\footnotesize liked};

\def \xpos {6.7}
\node [] (token) at ({\xpos * \xwarp},1) [text=gray] {\footnotesize liked};
\node [] (action) at ({\xpos * \xwarp}, 0) {\footnotesize $\textsc{shift}$};

\def \xpos {7.6}
\node [] (token) at ({\xpos * \xwarp},1) {\footnotesize their};
\node [] (action) at ({\xpos * \xwarp}, 0) {\footnotesize $\textsc{shift}$};

\def \xpos {8.6}
\node [] (action) at ({\xpos * \xwarp}, 0.5) {\footnotesize 10};
\node [] (action) at ({\xpos * \xwarp}, 0) {\footnotesize city};
\node [draw,rounded corners,dashed,thick] (boston) at ({\xpos * \xwarp}, 1) {\footnotesize Boston};
\node [draw,rounded corners,thick] (name) at ({\xpos * \xwarp}, 2) {\footnotesize name};
\draw [latex-] (boston) to node[right] {{\footnotesize op1}} (name);
\node [draw,rounded corners,thick] (city) at ({\xpos * \xwarp}, 3) {\footnotesize city};
\draw [latex-] (name) to node[right] {{\footnotesize name}} (city);

\def \xpos {9.7}
\node [] (action) at ({\xpos * \xwarp}, 0.5) {\footnotesize 11};
\node [] (action) at ({\xpos * \xwarp}, 0) {\footnotesize name};
\node [] (token) at ({\xpos * \xwarp},1) [text=gray] {\footnotesize Boston};
\node [] (arc\xpos) at ({\xpos * \xwarp}, -0.5) {\footnotesize $\textsc{ra(10,\textup{name})}$};

\def \xpos {11.3}
\node [] (action) at ({\xpos * \xwarp}, 0.5) {\footnotesize 13};
\node [] (action) at ({\xpos * \xwarp}, 0) {\footnotesize $\textsc{copy}$};
\node [] (token) at ({\xpos * \xwarp},1) [text=gray] {\footnotesize Boston};
\node [] (arc\xpos) at ({\xpos * \xwarp}, -0.5) {\footnotesize $\textsc{ra(11,\textup{op1})}$};

\def \xpos {12.3}
\node [] (token) at ({\xpos * \xwarp},1) [text=gray] {\footnotesize Boston};
\node [] (action) at ({\xpos * \xwarp}, 0) {\footnotesize $\textsc{shift}$};

\def \xpos {13.3}
\node [] (action) at ({\xpos * \xwarp}, 0.5) {\footnotesize 16};
\node [] (action) at ({\xpos * \xwarp}, 0) {\footnotesize trip-03};
\node [] (token) at ({\xpos * \xwarp},1) {\footnotesize trip};
\node [draw,rounded corners,thick] (trip03) at ({\xpos * \xwarp}, 3.5) {\footnotesize trip-03};
\path (token) edge[-,draw,dashed,thick] (trip03);
\draw [latex-] (person) to node[right] {{\footnotesize \textsc{arg0}}} (trip03);
\draw [latex-] (city) to node[right] {{\footnotesize \textsc{arg1}}} (trip03);
\draw [-latex] (like01) to node[right,bend left=30] {{\footnotesize \textsc{arg1}}} (trip03);
\node [] (arc\xpos) at ({\xpos * \xwarp}, -1) {\footnotesize $\textsc{ra(5,arg1)}$};
\node [] (arc\xpos) at ({\xpos * \xwarp}, -1.5) {\footnotesize $\textsc{la(1,arg0)}$};
\node [] (arc\xpos) at ({\xpos * \xwarp}, -0.5) {\footnotesize $\textsc{la(10,arg1)}$};

\def \xpos {14.3}
\node [] (action) at ({\xpos * \xwarp}, 0) {\footnotesize $\textsc{shift}$};
\node [] (token) at ({\xpos * \xwarp},1) [text=gray] {\footnotesize trip};

\end{tikzpicture}

\vspace{-2mm}
\caption{From top to bottom: graph (solid lines), sentence (source), addressable action positions and action sequence (target) for the sentence \textit{Employees liked their Boston trip}, aligned (dotted lines) to its AMR graph. Arc-creating actions are displayed vertically due to space constraints. Words are repeated in grey to indicate the word under cursor for each action. The node \textit{Boston} in dotted box is created by copying the token under cursor via $\cp$ action at position 13. $\textsc{la(1,arg0)}$ creates a left arc with label \textsc{arg0} from the top concept \textit{like-01} to the concept \textit{person} at position 1. For the concept \textit{trip-03}, $\textsc{la(1,arg0)}$ is a co-reference (re-entrancy) to the concept \textit{person}. 
}
    \label{fig:graph_orders}
\end{figure*}

\section{A Simplified Transition System}

In this section we propose a transition system for AMR parsing designed with two objectives: maximize the use of strong pre-trained decoders such as BART, and minimize the complexity and dependencies of the transition system compared to previous approaches. 
Similarly to \citet{zhou2021amr}, we scan the sentence from left to right and use a \textit{token cursor} to point to a source token at each step. Parser actions can either shift the cursor one token forward or generate any number of nodes and edges while the cursor points to the same token. 
The proposed set of actions is as follows:

\paragraph{$\shift$} moves token cursor one word to the right. 

\paragraph{\textrm{<string>}} creates node of name \textrm{<string>}. 

\paragraph{$\cp$} creates node where the node name is the token under the current cursor position.

\paragraph{$\la{\textup{j},\textsc{lbl}}$} creates an arc with label $\textsc{lbl}$ from the last generated node to the node generated at the $j_{th}$ transition step. 

\paragraph{$\ra{\textup{j},\textsc{lbl}}$} same as $\textsc{la}$ but with arc direction reversed.

\paragraph{\textsc{root}} declares the last predicted node as the root.\\

Unlike previous transition-based approaches, we do not use a reserved action, such as $\textsc{pred}$ \cite{zhou2021amr} or $\textsc{confirm}$ \cite{ballesteros2017amr}, to predict nodes;\footnote{$\textsc{pred}$ wraps the node name inside the action as $\textsc{pred}(\textrm{<string>})$, and $\textsc{confirm}$ calls a subroutine to predict the AMR node and find the right propbank sense.} instead we directly use the node name \textrm{<string>} as the action symbol generating that node.
This opens the possibility of utilizing BART's target side pre-trained vocabulary. We avoid using any copy actions that involve copying from lemmatizer outputs or lookup tables. Our $\textsc{copy}$ action is limited to copying the lower cased word. We also eliminate the use of $\textsc{subgraph}$ \cite{zhou2021amr} or $\textsc{entity}$ \cite{ballesteros2017amr} actions producing multiple connected nodes simultaneously, as well as $\textsc{merge}$ action creating spans of words. In previous approaches these actions were derived from alignments or hand-crafted. They thus did not cover all possible cases limiting the scalability of the approach. Finally, we discard the $\textsc{reduce}$ action previously used to delete a source token. The effect can be achieved by simply using $\shift$ without performing any other action. Figure \ref{fig:graph_orders} shows an example sentence with an action sequence and the corresponding graph. This can be compared with the handling of verbalization and named entities in Figure \ref{fig:graph_embedding}.

To train a parser with a transition system, we need an action sequence for each training sentence that will produce the gold graph when executed. This action sequence then serves as the target for seq-to-seq models. A simple rule-based \textit{oracle} algorithm creates these ground-truth sequences given a sentence, its AMR graph and node-to-word alignments. At each step, the oracle tries the following possibilities in the order of listing and performs the first one that applies:

\begin{enumerate}
\item Create next gold arc between last created node and previously created nodes;
\item Create next gold node aligned to token under cursor;
\item If not at sentence end, $\shift$ source cursor;
\item Finish oracle.
\end{enumerate}

If possible, nodes are generated by $\cp$ and otherwise with \textrm{<string>} actions. Arcs are generated with $\textsc{la}$ and $\textsc{ra}$, connecting the nodes closer to the current node before the ones that are farther away. Note that the arcs are created by pointing to positions in the action history, where a graph node is represented by the action that creates it, following \citet{zhou2021amr}. Multiple nodes can be generated at a single source word before the cursor is moved by $\shift$. When multiple nodes are aligned to the same token, these nodes are generated in a predetermined topological order of graph traversal, interleaved by edge creation actions. \textsc{root} is applied as soon as the root node is generated.

The above oracle circumvents the problem of unattachable nodes by avoiding the use of subgraph actions. This implies that the oracle will always produce a unique action sequence that fully recovers the gold graph as long as every node in the graph is aligned to some token. To guarantee that all nodes are aligned, we improve upon the alignment system from \citet{naseem2019rewarding, astudillo2020transition}, which aligns a large majority, but not all AMR nodes.\footnotemark\footnotetext{Roughly 1\% of graphs contain about 1-2 unaligned nodes.} In order to do this, we apply a heuristic based on graph proximity to maintain local correspondences between graph nodes and sentence words. If a node is not aligned, we copy the alignment from its first child node, if existing, and otherwise the alignment from its first parent node. For example, in Figure~\ref{fig:graph_orders} if the node \textit{person} was not provided with an alignment, our oracle would have assigned it to the aligned token of its child node \textit{employ-01}. This is a recursive procedure -- as long as there are some alignments to start with and the ground-truth graph is connected, all the nodes will get aligned.


Our proposed transition system makes better use of BART pre-trained decoder compared to previous transition-based approaches (see Section \ref{sec:results}) while greatly simplifying the transition set. It also naturally produces node-to-word alignments via source token cursor in the meantime.

\section{Parsing Model}
\label{sec:model}

We build our model on top of the pre-trained seq-to-seq Transformer, BART \citep{lewis2019bart}. We modify its architecture to incorporate a pointer network and internalize parser states induced by our transition system, and fine-tune for sentence-to-action generation.

\subsection{Structure-aware Architecture}

We adopt similar modifications on the Transformer architecture as in \citet{zhou2021amr} since our transition system is based on the same action-pointer mechanism.
The modifications do not introduce new modules or extra parameters, which naturally fit our need to adapt BART into a transition-based parser with internal graph well-formedness.

In particular, the target actions are factored into two parts: bare action symbols (containing labels when presented) and pointer values for edges. We use the BART standard output for the former, and a pointer network for the latter. As the pointing happens on the actions history, essentially a self-attention mechanism, we re-purpose one decoder self-attention head as the pointer network. It is supervised with additional cross entropy loss during fine-tuning and decoded for building graph edges at inference.

We encode the monotonic action-source alignments induced by the parser state with hard attention, i.e. by masking some decoder cross-attention heads to only focus on aligned words. Since BART processes source sentences with subwords, we apply an additional average pooling layer on top of its encoder to return states of original source words, used for the decoder layers for our hard attention.
At last, as the possible valid actions are constrained with transition rules and states at every step, we restrict the decoder output space via hard masking of the BART final softmax layer.
For simplicity, we do not incorporate the GNN-style \citep{li2019hierarchy} step-wise decoder graph embedding technique in \citet{zhou2021amr} as their gain was shown to be modest.

\subsection{Action Generation}



According to how we treat the target-side vocabulary for action generation, we propose two variations of the model. The first one is to use a completely separate vocabulary for target actions, where the decoder input side and output side use stand-alone embeddings for actions, separate from the pre-trained BART subword embeddings.\footnote{In practice the separate embeddings are initialized with the average subword embeddings from the original BART vocabulary, which gave small gains over random initialization.}
We denote this setup as our \textit{sep-voc} model (abbreviated as StructBART-S).

However, this might not fully utilize the power of the pre-trained BART since it is an encoder-decoder model with a single vocabulary and all embeddings shared. Although our generation targets are action symbols, the node generating actions are closely related to natural words in their surface forms, which are what BART was pre-trained on. Therefore, we propose a second variation where we use a joint vocabulary for both the source tokens and target actions. Naively relying on the original BART subword vocabulary would end up splitting action symbols blindly, which is not desired as the structures such as alignments and edge pointers would be disrupted. Inspired by \citet{bevilacqua2021one}, we add frequent node-creating actions to the vocabulary, in order to capture common AMR concepts intact, and split the remaining concepts with BART subword vocabulary. Non-node-creating actions such as $\shift$ and $\cp$ are added as-is to the BART vocabulary.

In this setup, a single node string can potentially be generated with multiple steps; we modify the arc transitions to always point to the beginning position of a node string for attachment. With joint vocabulary setup, the model could learn to generate unseen nodes with BART's subword vocabulary, eliminating potential out-of-vocabulary problems. We refer to this setup as our \textit{joint-voc} model (abbreviated as StructBART-J).

\subsection{Training and Inference}

We load the pre-trained BART parameters except for the standalone vocabulary embeddings for sep-voc model and the extended embeddings for the joint-voc model. 
We then fine-tune the model with the updated structure-aware architectures on sentence-action pairs with addition of pointer loss.

For decoding, we use similar constrained beam search algorithm as in \citet{zhou2021amr}, but with our own transition set and rules. We run a state machine on the side to get parser states used by the model.
Note that for our joint-voc model, we only allow subword split for node (\textrm{<string>}) actions.
As our fine-tuned model is already structure-aware, the graph well-formedness is always guaranteed and no post-processing is needed to return valid graphs, unlike \citet{xu2020improving, bevilacqua2021one}.
The only post-processing we use is to add wikification nodes as used in all previous parsers.\footnote{We also do light cleaning of the decoded AMR when they are printed to penman format, such as removing reserved characters in node concepts and printing disconnected subgraphs.}

\section{Experimental Setup}
\label{sec:exp_setup}

\paragraph{Datasets}

We evaluate our models on 3 AMR benchmark datasets, namely AMR 1.0 (LDC2014T12), AMR 2.0 (LDC2017T10), and AMR 3.0 (LDC2020T02). They have around 10K, 37K, and 56K sentence-AMR pairs for training, respectively.\footnote{See Appendix \ref{sec:appendix_data} for detailed dataset sizes. Data source: \url{https://amr.isi.edu/download.html}.}
Both AMR 2.0 and AMR 3.0 have wikification nodes but AMR 1.0 does not.


\paragraph{Evaluation}

We assess our models with $\smatch$ (F1) scores\footnote{\url{https://github.com/snowblink14/smatch/tree/v1.0.4}.} \citep{cai2013smatch}. We also report the fine-grained evaluation metrics \citep{damonte2016incremental} to further investigate different aspects of parsing results, such as concept identification, entity recognition, re-entrancies, etc.



\begin{table*}[!t]
    \centering
    \resizebox{\textwidth}{!}{%
    \begin{tabular}{c|l|c|c|ccc|c|c|ccc}
    \toprule
        \multirow{2}{*}{ID} & \multirow{2}{*}{Model} & \multirow{2}{*}{\begin{tabular}{@{}c@{}} Pre-trained \\ Model \end{tabular}} & \multirow{2}{*}{\begin{tabular}{@{}c@{}} Collapse \\ Subgraph \end{tabular}} & \multicolumn{3}{c}{External Dependency} & \multirow{2}{*}{\begin{tabular}{@{}c@{}} Extra \\ Data \end{tabular}} & \multirow{2}{*}{\begin{tabular}{@{}c@{}} Train \\ Align. \end{tabular}} & \multicolumn{3}{c}{$\smatch$ F1 (\%)} \\
        & & & & POS & NER & Lemm. & & & AMR 1.0 & AMR 2.0 & AMR 3.0 \\
    \midrule
       1 & \citet{naseem2019rewarding} & BERT & S.A. & & & \ding{51} & & \ding{51} & - & 75.5 & - \\
       2 & \citet{zhang2019amr} & BERT & G.R.& \ding{51} & \ding{51} & & & & 70.2 \small{$\pm$0.1} & 76.3 \small{$\pm$0.1} & - \\
       3 & \citet{zhang2019broad} & BERT & G.R.& \ding{51} & \ding{51} & & & & 71.3 \small{$\pm$0.1} & 77.0 \small{$\pm$0.1} & -\\
       4 & \citet{cai2020amr} & BERT & & \ding{51} & \ding{51} & \ding{51} & & &  74.0 & 78.7 & - \\
       5 & \citet{cai2020amr} & BERT & G.R. & \ding{51} & \ding{51} & \ding{51} & & &  75.4 & 80.2 & -\\
       6 & \citet{astudillo2020transition} & RoBERTa & S.A. & & & \ding{51} & & \ding{51} &  76.9 \small{$\pm$0.1} & 80.2 \small{$\pm$0.0} & - \\
       7 & \citet{lyu2020differentiable} & RoBERTa & G.R. & & \ding{51} & \ding{51} & & & - & - & 75.8 \\
       8 & \citet{bevilacqua2021one} & BART$^\text{\ding{61}}$ & & & & & & & - & 83.8 & 83.0 \\
       9 & \citet{bevilacqua2021one} & BART$^\text{\ding{61}}$ & G.R. & & \ding{51} & & & & - & 84.5 & 80.2 \\
       10 & \citet{zhou2021amr} & RoBERTa & S.A. & & & \ding{51} & & \ding{51} & 78.3 \small{$\pm$0.1} & 81.7 \small{$\pm 0.1$} & 80.3 \small{$\pm$0.1} \\
    \midrule
       11 & \citet{xu2020improving} & Custom$^\text{\ding{61}}$ & & & & & 4M & & - & 81.4 & - \\
       12 &\citet{lee2020pushing} & RoBERTa & S.A. & & & \ding{51} & 85K & \ding{51} &  78.2 \small{$\pm$0.1} & 81.3 \small{$\pm$0.0} & - \\
       13 & \citet{bevilacqua2021one} & BART$^\text{\ding{61}}$ &  & & & & 200K & & - & 84.3 & \textbf{83.0} \\
       14 & \citet{zhou2021amr} & RoBERTa & S.A. & & & \ding{51} & 70K & \ding{51} & - & 82.6 \small{$\pm 0.1$} & - \\
    \midrule
       15 &StructBART-S & BART$^\text{\ding{61}}$ &  & & & &  & \ding{51} & 81.6 \small{$\pm 0.1$}  &  84.0 \small{$\pm 0.1$} & 82.3 \small{$\pm 0.0$}  \\
       16 & StructBART-J & BART$^\text{\ding{61}}$ &  & & & &  & \ding{51} & \textbf{81.7 \small{$\pm$0.2}} &  84.2 \small{$\pm 0.1$} & 82.0 \small{$\pm 0.0$} \\
       17 & StructBART-S & BART$^\text{\ding{61}}$ &  & & & & 47K & \ding{51} & - & - & 82.7 \small{$\pm0.1$} \\
       18 & StructBART-J & BART$^\text{\ding{61}}$ &  & & & & 47K & \ding{51} & - & \textbf{84.7 \small{$\pm$0.1}} & 82.6 \small{$\pm0.1$} \\
    \midrule
       19 & StructBART-S ensem. & BART$^\text{\ding{61}}$ &  & & & & 47K & \ding{51} & - & -  & \textbf{83.1} \\
       20 & StructBART-J ensem. & BART$^\text{\ding{61}}$ &  & & & & 47K & \ding{51} & - & \textbf{84.9}   & - \\
    \bottomrule
    \end{tabular}
    } 
    \caption{$\smatch$ (\%) scores on AMR 1.0, 2.0, and 3.0 test data, associated with each model's dependency on various resources. 1-10/11-14: previous models without/with extra data; 15-18: our models (-S/-J for separate/joint vocabularies for source and target); 19-20: our models with ensemble decoding. Symbols indicate: G.R. - graph re-categorization, S.A. - subgraph action used in transition-based parsers (both detailed in Section \ref{sec:complexity}), POS - part of speech tagger, NER - named entity recognizer, Lemm. - lemmatizer, Align. - alignments (only used at training time). $^\text{\ding{61}}$ indicates fine-tuning on top of pre-trained model. All models rely on a external wikification method (ommited). Our results are average of 3 runs with different random seeds. We also report standard deviation and provide ensemble results for the 3 seed combination.} 
    \label{tab:main_results123}
\end{table*}

\begin{table*}[!t]
    \centering
    \resizebox{\textwidth}{!}{%
    \begin{tabular}{c|l|c|cccccccc}
        \toprule
        ID & Model & \textsc{Smatch} & Unlabel & NoWSD & Concepts & NER & Negation & Wiki. & Re-entrancy & SRL \\
        \midrule
        1 & \citet{naseem2019rewarding} & 75.5 & 80 & 76 & 86 & 83 & 67 & 80 & 56 & 72 \\
        4 & \citet{cai2020amr} & 78.7 & 81.5 & 79.2 & 88.1 & 87.1 & 66.1 & 81.3 & 63.8 & 74.5 \\
        6 & \citet{astudillo2020transition} & 80.2 & 84.2 & 80.7 & 88.1 & 87.5 & 64.5 & 78.8 & 70.3 & 78.2 \\
        8 & \citet{bevilacqua2021one} & 83.8 & 86.1 & 84.4 & 90.2 & 90.6 & \textbf{74.4} & \textbf{84.4} & 70.8 & 79.6 \\
        10 & \citet{zhou2021amr} & 81.8 & 85.5 & 82.3 & 88.7 & 88.5 & 69.7 & 78.8 & 71.1 & 80.8 \\
        \midrule 
        15 & StructBART-S & 84.1 & 87.5 & 84.4 & 90.4 & \textbf{92.2} & 71.0 & 79.6 & 73.9 & 83.0 \\
        16 & StructBART-J & \textbf{84.3} & \textbf{87.9} & \textbf{84.7} & \textbf{90.6} & 92.1 & 72.5 & 80.8 & \textbf{74.3} & \textbf{83.4} \\
        \bottomrule
    \end{tabular}
    } 
    \caption{Fine-grained F1 scores on the AMR 2.0 test set, among models that do not use extra silver data and graph re-categorization.
    The model IDs are matched with those in Table~\ref{tab:main_results123} for detailed model features.
    We report results with our single best model (selected on development data) for fair comparison.}
    \label{tab:finegrained_amr2.0}
\end{table*}

\paragraph{Model Configuration}

We follow the original BART configuration \cite{lewis2019bart} and code.
\footnote{\url{https://github.com/pytorch/fairseq/tree/v0.10.2/examples/bart}.}
We use the large model configuration as default, and also the base model for ablation studies.
The pointer network is always tied with one head of the decoder top layer, and the pointer loss is added to the model cross-entropy loss with 1:1 ratio for training. Transition alignments are used to mask cross-attention in 2 heads of all decoder layers.
For sep-voc model, we build separate embedding matrices for target actions from the training data for decoder input and output space.
For joint-voc model, we add new embedding vectors for non-node action symbols and node action strings with a default minimum frequency of 5 (only accounts for about one third of all nodes due to sparsity).

\paragraph{Implementation Details}

Our models are trained with Adam optimizer with batch size 2048 tokens and gradient accumulation of 4 steps. Learning rate is $1\mathrm{e}{-4}$ with 4000 warm-up steps using the inverse-sqrt scheduling scheme \citep{vaswani2017attention}.
The hyper-parameters are fixed and not tuned for different models and datasets, as we found results are not sensitive within small ranges. 
We train sep-voc models for 100 epochs and joint-voc models for 40 epochs as the latter is found to converge faster.
The best 5 checkpoints based on development set $\smatch$ from greedy decoding are averaged, and default beam size of 10 is used for decoding for our final parsing scores.
We implement our model\footnote{Code and model available at \url{https://github.com/IBM/transition-amr-parser}.} with the \textsc{fairseq} toolkit \citep{ott2019fairseq}.
More details can be found in the Appendix.


\section{Results}
\label{sec:results}

\begin{table*}
    \centering
    \resizebox{\textwidth}{!}{%
    \begin{tabular}{cccccccc}
        \toprule
            \multirow{3}{*}{Transition System} & \multicolumn{3}{c}{Features}  & \multicolumn{2}{c}{Model Results on AMR 2.0} & \multicolumn{2}{c}{Model Results on AMR 3.0} \\
            \cmidrule(lr){2-4}
            \cmidrule(lr){5-6}
            \cmidrule(lr){7-8}
            & \multirow{2}{*}{\begin{tabular}{@{}c@{}} \#Base \\ Actions \end{tabular}} & \multirow{2}{*}{\begin{tabular}{@{}c@{}} Distant \\ Edges \end{tabular}} & \multirow{2}{*}{\begin{tabular}{@{}c@{}} Special \\ Subgraph \end{tabular}} &  \multirow{2}{*}{\begin{tabular}{@{}c@{}} APT$^*$ \\ \citep{zhou2021amr} \end{tabular}  } & \multirow{2}{*}{\begin{tabular}{@{}c@{}} StructBART \\ sep-voc \end{tabular} } &
            \multirow{2}{*}{\begin{tabular}{@{}c@{}} APT$^*$ \\ \citep{zhou2021amr} \end{tabular}  } & \multirow{2}{*}{\begin{tabular}{@{}c@{}} StructBART \\ sep-voc \end{tabular} } \\
            & & & & & & & \\
        \midrule
            \citet{astudillo2020transition} & 12 & \textsc{Swap} & merge  & - & - & - & -\\
            \citet{zhou2021amr} & 10 & pointer & merge  & 81.5 \small{$\pm$0.1} & 83.4 \small{$\pm$0.1} & 79.8 \small{$\pm$0.1} & 81.6 \small{$\pm$0.0} \\
            Ours & 6 & pointer & no & 81.6 \small{$\pm$0.1} & 84.0 \small{$\pm$0.1} & 79.6 \small{$\pm$0.0} & 82.3 \small{$\pm$0.1} \\
        \bottomrule
    \end{tabular}
    } 
    \caption{Transition system comparison, including their effects on different parsing models.
    $^*$ we adopt the cited model without graph structure embedding to compare and run on our proposed oracle.}
    \label{tab:oracle_compare_wide_full}
\end{table*}


\begin{table}[!t]
    \centering
    \resizebox{\columnwidth}{!}{%
    \begin{tabular}{ccc}
    \toprule
        Transition System & Avg. \#actions & Oracle $\smatch$ \\
    \midrule
        \citet{astudillo2020transition} & 76.2 & 98.0 \\
        \citet{zhou2021amr} & 41.6 & 98.9 \\
        Ours & 45.6 & 99.9 \\
    \bottomrule
    \end{tabular}
    } 
    \caption{Average action sequence length and oracle coverage on AMR 2.0 training data from different transition systems.
    Average source sentence length is 18.9.
    }
    \label{tab:oracle_actions}
\end{table}

\paragraph{Main Results}

We present parsing performances of our model (StructBART) in comparison with previous approaches in Table~\ref{tab:main_results123}. For each model, we also list its features such as utilization of pre-trained language models and graph simplification methods such as re-categorization. This gives a comprehensive overview of how systems compare in terms of complexity aside from performance.

All recent systems rely on pre-trained language models, either as fixed features or through fine-tuning. 
The pre-trained BART is particularly beneficial due to its encoder-decoder structure. 
Among all the models, the graph linearization models \citep{xu2020improving, bevilacqua2021one} have the least number of extra dependencies when not using graph re-categorization. Our model only requires aligned training data, a trait common to all transition-based approaches. This bears the advantage of producing reliable alignments at decoding time, which are useful for downstream tasks and as explanation of the graph constructing process. 


Both our sep-voc and joint-voc model variations work well on all datasets.
Without using extra silver data, our model achieves the $\smatch$ score of 84.2 {\small $\pm0.1$} on AMR 2.0, which is the same as the previous best model \citep{bevilacqua2021one} with 200K silver data. With the input of only 47K silver data (consisting of $\sim$20K example sentences of propbank frames and randomly selected $\sim$27K SQuAD-2.0 context sentences\footnote{https://rajpurkar.github.io/SQuAD-explorer/.}), we achieve the highest score of 84.7 {\small $\pm0.1$} for AMR 2.0. We also attain the high score of 81.7 {\small $\pm0.2$} on the smallest AMR 1.0 benchmark, and the second best score of 82.7 {\small $\pm0.1$} on the largest AMR 3.0 benchmark. Ensemble of the 3 models from the silver training further improves the performances to 84.9 for AMR 2.0 and 83.1 for AMR 3.0.

\paragraph{Fine-grained Results}

We further examine the fine-grained parsing results on AMR 2.0 in Table~\ref{tab:finegrained_amr2.0}. We compare models not relying on extra data nor graph re-categorizationn since silver data sets differ across methods, and re-categorization comes with limitations outlined in Section \ref{sec:complexity}. Our models achieve the highest scores across most of the categories, except for negation and wikification. The former may be due to alignment errors and the latter is solved as a separate post-processing step independent of the parser. Compared with the closely related model from \citet{bevilacqua2021one} which also fine-tunes BART but directly on linearized graphs, we achieve significant gains on re-entrancy and SRL ($\textsc{:arg-i}$ arcs), proving our
model generate AMR graphs more faithful to their topological structures.

\section{Analysis}
\label{sec:analysis}

\paragraph{Transition System}

Table~\ref{tab:oracle_compare_wide_full} and Table~\ref{tab:oracle_actions} compare different transition systems used by recent transition-based AMR parsers with strong performances. Our proposed system has the smallest set of base actions, utilizes the action-side pointer mechanism for flexible edge creation as in \citet{zhou2021amr}, but does not rely on special treatment of certain subgraphs such as named entities and dates. This results in slightly longer action sequences compared to \citet{zhou2021amr}, but with almost 100\% coverage\footnotemark\footnotetext{We can recover 100\% of AMR 2.0 training graphs excluding $4$ with notation errors. Imperfect $\smatch$ is due to ambiguities of our parser in recovering Penman notation.} (Table~\ref{tab:oracle_actions}).
Our transition system and oracle can always find action sequences with full recovery of the original AMR graph, regardless of graph topology and alignments.

To assess whether our proposed transition system helps integration with pre-trained BART,
we train both the APT model from \citet{zhou2021amr} and our sep-voc model on the transition system of \citet{zhou2021amr} and the one introduced in this work (Table~\ref{tab:oracle_compare_wide_full} last 4 columns). The APT model, based on fixed RoBERTa features, does not benefit from the proposed transition system. However, our proposed model gains 0.6 on AMR 2.0 and 0.7 on AMR 3.0. This confirms the hypothesis that the proposed transitions are better able to exploit BART's powerful language generation ability.

\begin{table}[!t]
    \centering
    \resizebox{\columnwidth}{!}{%
    \begin{tabular}{ccc}
    \toprule
         \multirow{2}{*}{\begin{tabular}{@{}c@{}} Model \\ Variation \end{tabular}} & \multicolumn{2}{c}{
        $\smatch$ (\%)} \\
        & sep-voc & joint-voc \\
    \midrule
         Ours (hard attention) & 84.0 \small{$\pm$0.1} & 84.2 \small{$\pm$0.1} \\
    \midrule
         No align. modeling  & 83.5 \small{$\pm$0.0} & 83.4 \small{$\pm$0.2} \\
         Align. soft supervision  & 82.9 \small{$\pm$0.0} & 83.0 \small{$\pm$0.0} \\
         Align. add src emb.  & 83.9 \small{$\pm$0.0} & 84.1 \small{$\pm$0.0} \\
         No $\cp$ action  & 83.1 \small{$\pm$0.1} & 84.1 \small{$\pm$0.0} \\
    \bottomrule
    \end{tabular}
    } 
    \caption{Ablation study of structure modeling with transition alignments.
    Results are on AMR 2.0 test data.}
    \label{tab:structure_ablation_compact}
\end{table}

\paragraph{Structural Alignment Modeling}

In Table~\ref{tab:structure_ablation_compact}, we evaluate the effects of our structural modeling of parser states within BART during fine-tuning.
 Action-source alignments are natural byproduct of the parser state, providing structural information of where and how to generate the next graph component.
 Our default use of hard attention to encode such alignments works the best.
We explore two other strategies for modeling alignments. One is to supervise cross-attention distributions for the same heads with inferred alignments during training, inspired by \citet{strubell2018linguistically}. The other is to directly add the aligned source contextual embeddings from the encoder top layer to the decoder input at every generation step. The former hurts the model performance, indicating the model is unable to learn the underlying transition logic to infer correct alignments, while the latter does equally well as our default model. These results justify the modeling of structural constraints, even when fine-tuning strong pre-trained models such as BART.
 
We also ablate the use of $\cp$ action in our transition system. The sep-voc model suffers but the joint-voc model is not affected. Without $\cp$ action, the joint-voc model would rely more on BART's pre-trained subword embeddings to split node concepts more frequently, while the sep-vocab model would need to learn to generate more rare concepts from scratch. This indicates that BART's strong generation power is fully used to tackle concept sparsity problems with its subwords.

\begin{figure}
    \centering
    \includegraphics[width=0.99\linewidth]{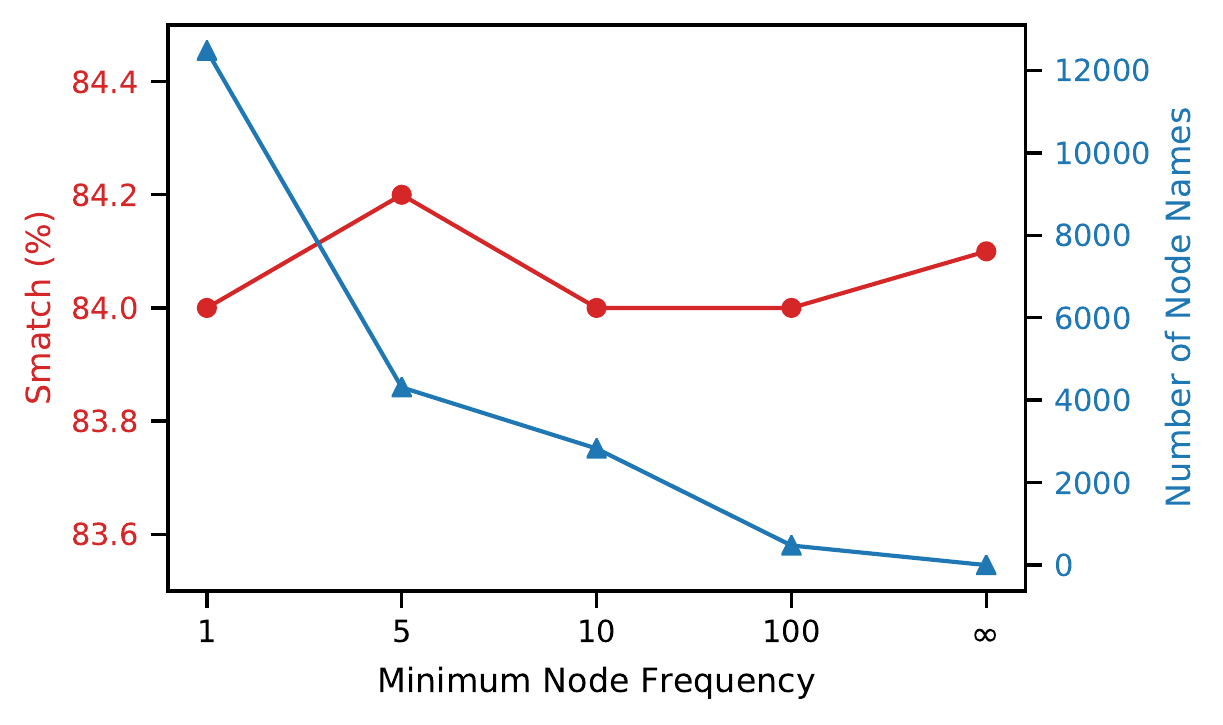}
    \caption{Effect of special AMR node names added to the BART vocabulary in joint-voc model on AMR 2.0 dataset. Remaining AMR concepts are split and generated with BART subwords and sense numbers.}
    \label{fig:voc_min_node}
\end{figure}


\begin{table}[!t]
    \centering
    \resizebox{\columnwidth}{!}{%
    \begin{tabular}{c|cccccc}
    \toprule
      \multirow{2}{*}{\#} & \multicolumn{4}{c}{Model Initialization}  &  \multicolumn{2}{c}{$\smatch$ (\%)} \\
    & src emb & encoder & decoder & tgt emb & Base & Large \\
    \midrule
       1 & & & & & 71.2 & 72.7 \\
       2 & \ding{51} & & & \ding{51} & 71.7  & 72.8 \\
       3 & \ding{51} & \ding{51} & & \ding{51} & 81.4 & 82.8 \\
       4 & \ding{51} &  & \ding{51} & \ding{51} & 69.2 & 9.5$^*$ \\
       5 & \ding{51} &  &  &  & 71.2 & 72.8 \\
       6 & \ding{51} & \ding{51} &  &  & 80.9 & 82.5 \\
       7 & \ding{51} & \ding{51} & \ding{51} &  & 82.2  & 83.9 \\
       8 & \ding{51} & \ding{51} & \ding{51} & \ding{51} & 82.7 & 84.1 \\
    \midrule
       9 & \multicolumn{4}{c}{freeze BART src emb} & 82.6 & 84.0 \\
       10 & \multicolumn{4}{c}{freeze BART src emb + encoder} & 80.9 & 81.8 \\
    \bottomrule
    \end{tabular}
    } 
    \caption{Effects of pre-trained BART parameters. Results are with our sep-voc model on AMR 2.0 data. $^*$ failed to converge with a range of hyper-parameters.}
    \label{tab:bart_components}
\end{table}

\paragraph{Special Nodes in Joint-Voc}

In Figure~\ref{fig:voc_min_node},
we show the joint-voc model performance with different sized (joint) vocabularies. The vocabulary size is controlled by specifying the minimum frequency of occurrence needed for an AMR concept to be added to the vocabulary.
For instance, when the minimum frequency is 1, all 12475 AMR concepts from the training data are added onto the BART vocabulary. The number of added concepts decreases as we increase the minimum  frequency threshold. On model performance side, we only observe $\pm$0.2 score variations resulting from vocabulary expansion. More interestingly, the model can work equally well when no special concepts are added to the BART vocabulary (minimum node frequency is $\infty$) -- where all the node names are split and generated with BART subword tokens.
Although our default setup uses frequency threshold of 5 in joint-voc expansion, following \citet{bevilacqua2021one}, it seems unnecessary in terms of achieving good performance.
This highlights the efficacy of utilizing the pre-trained BART's language generation power for AMR concepts even with relatively small annotated training datasets.

\paragraph{Pre-trained Parameters}

We study the contribution of different pre-trained BART components in Table~\ref{tab:bart_components}. With our sep-voc model, we decompose the whole seq-to-seq Transformer into four components for BART initialization, i.e. the source embedding (mapped with BART shared embedding), encoder, decoder, and the separate target embedding (initialized with the average subword embeddings from BART shared embedding). We run both the StructBART base model and the StructBART large model with different combinations of parameter initialization, on the top part of Table~\ref{tab:bart_components}.
We can see that a randomly initialized model of the same size (\#1) performs badly.
There is an accumulative effect of BART initialization in helping the model performance, except that BART decoder can not work alone well without its encoder (\#4).
The encoder gives the largest performance gains (\#3 vs. \#2, \#6 vs. \#5) of about 10 points.
Adding the decoder further gives around 1.4 points on top (\#7 vs. \#6), justifying its importance as well.

We also experiment with freezing BART parameters during training in the bottom part of Table~\ref{tab:bart_components}.
Our results of freezing the BART encoder are on similar levels of previous best RoBERTa feature based models, which is behind the full fine-tuning.
Overall, full initialization from BART with structure-aware fine-tuning (\#8) works the best.

\section{Related Work}
\label{sec:related_work}


Using seq-to-seq to predict linearized graph sequences for parsing was proposed in \citet{vinyals2015grammar} and is currently a very extended approach \cite{van2017neural, ge2019modeling, rongali2020don}. However, it is only recently with the rise of pre-trained Transformer decoders, that these techniques have become dominant in semantic parsing. \citet{xu2020improving} proposed custom multi-task pre-training and fine-tuning approach for conventional Transformer models \cite{vaswani2017attention}. The massively pre-trained transformer BART \cite{lewis2019bart} was used for executable semantic parsing in \citet{chen2020low} and AMR parsing in \citet{bevilacqua2021one}. The importance of strongly pre-trained decoders seems also justified as BART gains popularity in various semantic generation tasks \citep{chen2020low, shi2020learning}. Our work aims at capitalizing on the outstanding performance shown by BART, while providing a more structured approach that guarantees well-formed graphs and yields other desirable sub-products such as alignments. We show that this is not only possible but also attains state-of-the art parsing results without graph re-categorization.
Our analysis also shows that contrary to \citet{xu2020improving}, vocabulary sharing is not necessary for strong performance for our structural fine-tuning.

Encoding of the parser state into neural parsers has been undertaken in various works, including seq-to-seq RNN models \citep{liu-zhang-2017-encoder,zhang-etal-2017-stack,buys2017robust}, encoder-only Transformers \citep{ahmad-etal-2019-difficulties}, seq-to-seq Transformers \cite{astudillo2020transition,zhou2021amr} and pre-trained language models \cite{qian-etal-2021-structural}. Here we explore the application of these approaches to pre-trained seq-to-seq Transformers. Borrowing ideas from \citet{zhou2021amr}, we encode alignment states into the pre-trained BART attention mechanism, and re-purpose its self-attention as a pointer network. We also rely on a minimal set of actions targeted to utilize BART's generation with desirable guarantees, such as no unattachable nodes and full recovery of all graphs.
We are the first to explore transition-based parsing applied on fine-tuning strongly pre-trained seq-to-seq models, and we demonstrate that parser state encoding is still important for performance, even when implemented inside of a powerful pre-trained decoder such as BART.

\section{Conclusion}


We explore the integration of pre-trained sequence-to-sequence language models and transition-based approaches for AMR parsing, with the purpose of retaining the high performance of the former and structural advantages of the latter. We show that both approaches are complementary, establishing the new state of the art for AMR 2.0. Our results indicate that instead of simply converting the structured data into unstructured sequences to fit the need of the pre-trained model, it is possible to effectively re-purpose a generic pre-trained model to a structure-aware one achieving strong performance.
Similar principles can be applied to adapt other powerful pre-trained models such as T5 \citep{raffel2019exploring} and GPT-2 \citep{radford2019language} for structured data predictions. It is worth exploring thoroughly the pros and cons of introducing structure to the model compared to removing structure from the data (linearization) in various scenarios.

\bibliography{citation}
\bibliographystyle{acl_natbib}

\newpage

\appendix


\section{Dataset Statistics}
\label{sec:appendix_data}

We list the dataset sizes of AMR benchmarks in Table~\ref{tab:app_data}. The sizes increase with the release version number. AMR 2.0 is the most used by far. AMR 2.0 shares the same set of sentences for development and test data with AMR 1.0, but with revised annotations and wikification links. AMR 3.0 is released most recently, which is under-explored.

Our silver data originate from two sources. First, we use $\sim$20K example sentences ($\sim$386K tokens) from the propbank frames included in the AMR 3.0 distribution. Second, we use randomly selected $\sim$27K sentences ($\sim$620K tokens) from SQuAD 2.0 context sentences available from \url{https://rajpurkar.github.io/SQuAD-explorer/}.

\begin{table}[h]
    \centering
    \resizebox{\columnwidth}{!}{%
    \begin{tabular}{cccc}
    \toprule
        Data Split & AMR 1.0 & AMR 2.0 & AMR 3.0  \\
    \midrule
        Training & 10,312 & 36,521 & 55,635 \\
        Development & 1,368 & 1,368 & 1,722 \\
        Test & 1.371 & 1.371 & 1,898 \\
    \bottomrule
    \end{tabular}
    } 
    \caption{Number of sentence-AMR instances in the AMR benchmark datasets.}
    \label{tab:app_data}
\end{table}

\section{Details of Model Structures and Number of Parameters}
\label{sec:appendix_param}

\begin{table}[!t]
    \centering
    \resizebox{\columnwidth}{!}{%
    \begin{tabular}{ccc}
    \toprule
        Configuration & BART Base & BART Large \\
    \midrule
        Encoder layers & 6 & 12 \\
        Heads per layer & 12 & 16 \\
        Hidden size & 768 & 1024 \\
        FFN size & 3072 & 4096 \\
        Size of vocab & 51201$^*$ & 50265 \\
    \midrule
        Size of emb. matrix & 39,322,368 & 51,471,360 \\
        \#parameters trained & 140,139,266 & 406,291,458 \\
    \bottomrule
    \end{tabular}
    }  
    \caption{Original model configurations of pre-trained BART from \textsc{fairseq} (\url{https://github.com/pytorch/fairseq/tree/v0.10.2/examples/bart}). The embeddings for source, decoder input and output are all shared and thus the same (not counted as extra in training parameters). $^*$vocabulary for the base model is larger due to additional paddings at the end, but effective vocabulary symbols are the same as the large model.}
    \label{tab:app_bartparam}
\end{table}

\begin{table}[!t]
    \centering
    \resizebox{\columnwidth}{!}{%
    \begin{tabular}{ccccc}
    \toprule
        Model & Param. &  AMR 1.0 & AMR 2.0 & AMR 3.0 \\
    \midrule
         \multirow{3}{*}{sep-voc} & Src vocab size & 50265 & 50265 & 50265 \\
          & Tgt vocab size & 6976 & 12752 & 16180 \\
         & \#param. trained & 420,578,304 & 432,407,552 & 439,436,288 \\
    \midrule
         \multirow{2}{*}{joint-voc} & joint vocab size & 51921 & 53487 & 54388 \\
         & \#param. trained & 407,987,200 & 409,590,784 & 410,517,504 \\
    \bottomrule
    \end{tabular}
    }  
    \caption{Model parameters of our StructBART (large model). The sizes differ based on the target side vocabulary, which is dependent on different training data. Addition of silver training data adds only a fraction of the parameters to the benchmark datasets.}
    \label{tab:app_modelparam}
\end{table}

In Table~\ref{tab:app_bartparam}, we list the detailed model configuration and number of parameters of the official pre-trained BART models.
Our fine-tuned StructBART is with different action vocabulary strategies which builds additional embedding vectors for certain action symbols. The numbers vary from training dataset. We list the detailed number of parameters of our fine-tuned model in Table~\ref{tab:app_modelparam}. The fine-tuned model only increases about 3\%-8\% more parameters for sep-voc model and 0.4\%-1\% more parameters for joint-voc model.

\section{Implementation Details}
\label{sec:appendix_implementation}

We use the Adam optimizer with $\beta_1=0.9$ and $\beta_2=0.98$. Batch size is set to 2048 maximum number of tokens, and gradient is accumulated over 4 steps.
The the learning rate schedule is the same as \citet{vaswani2017attention}, where we use the maximum learning rate of $1\mathrm{e}{-4}$ with 4000 warm-up steps.
Dropout of rate 0.2 and label smoothing of rate 0.01 are used.
These hyper-parameters are fixed and not tuned for different models and datasets, as we found results are not sensitive within small ranges.
Without silver data, we train sep-voc models for 100 (AMR 1.0 \& 2.0) or 120 (AMR 3.0) epochs and joint-voc models for 40 epochs as the latter is found to converge faster. The best 5 (AMR 1.0 \& 2.0) or 10 (AMR 3.0) checkpoints among the last 40/30 epochs are selected based on development set $\smatch$ from greedy decoding and averaged over the model parameters as our final model. With the 50K silver data, we train both sep-voc and joint-voc models for 20 epochs and select the best 10 checkpoints for model parameter averaging. We use a default beam size of 10 for decoding for our final parsing scores.
Our models are implemented with the \textsc{fairseq} toolkit \citep{ott2019fairseq}, trained and tested on a single Nvidia Tesla V100 GPU with 32GB memory.
We use fp16 mixed precision training whenever possible, with which training a large model on AMR 2.0 takes about 10 hours for sep-vocab models and 7 hours for joint-vocab models, and the time varies proportionally with data size for other datasets and with silver data.





\end{document}